# Detection of Homophobia & Transphobia in Dravidian Languages: *Exploring Deep Learning Methods*

Deepawali Sharma, Vedika Gupta &Vivek Kumar Singh

**Abstract.** The increase in abusive content on online social media platforms is impacting the social life of online users. Use of offensive and hate speech has been making social media toxic. Homophobia and transphobia constitute offensive comments against LGBT+ community. It becomes imperative to detect and handle these comments, to timely flag or issue a warning to users indulging in such behaviour. However, automated detection of such content is a challenging task, more so in Dravidian languages which are identified as low resource languages. Motivated by this, the paper attempts to explore applicability of different deep learning models for classification of the social media comments in Malayalam and Tamil languages as homophobic, transphobic and non-anti-LGBT+content. The popularly used deep learning models- Convolutional Neural Network (CNN), Long Short Term Memory (LSTM) using GloVe embedding and transformer-based learning models (Multilingual BERT and IndicBERT) are applied to the classification problem. Results obtained show that IndicBERT outperforms the other implemented models, with obtained weighted average F1-score of 0.86 and 0.77 for Malayalam and Tamil, respectively. Therefore, the present work confirms higher performance of IndicBERT on the given task in selected Dravidian languages.

## 1 Introduction

People share and post their views/opinions/thoughts about various topics on various social media platforms. The freedom to create such content also results in a large amount of displeasing data on racism, homophobia, transphobia, targeting particular communities and organizations, posted on the Web. Unfortunately, LGBT+ people are being abused online by other communities resulting into problems of mental disorders (anxiety disorders, behavioural and emotional disorders, depression) in them. Sometimes, campaigns are initiated against these vulnerable communities when they come online for support and these campaigns are often converted into bullying [8]. Automatic detection of such content is a challenging task, more so if the text is in a low resource language. Many studies have been done on the detection of hate speech [9][10][11], offensive and abusive content [12][13][14], cyberbullying [15][16][17] and many more on content in some selected languages. However, there is very little effort towards detecting racism, homophobia, transphobia in online texts from low resource Indian languages.

This paper presents an attempt towards this direction. Different deep learning-based models (CNN, LSTM, mBERT and IndicBERT) are implemented on data from two languages, namely Tamil and Malayalam. Both languages belong to the family of Dravidian languages and are low-resourced languages. The dataset provided by DravidianLangTech [8] is used for the purpose of the experiment. The comments are classified as Homophobic, Transphobic and Non-anti-LGBT+content. Those comments that contain pejorative labels or denigrative phrases against the gay, lesbian or any other community are classified as Homophobic [8]. People who can be homosexual, or can be transphobic without being homophobic, are identified as transphobic persons. Therefore, the comments that advocate or proposes anything unfavourable for transgender people are classified as transphobic [8]. Those comments that do not include homophobic, transphobic and have positive influence on readers are classified as non-anti-LGBT+content.

Different deep learning-based models are implemented to detect homophobia and transphobia in comments. More specifically, the paper performs the following tasks:

- Implementation of deep learning models (CNN, LSTM) with GloVe embedding for data in both languages (Malayalam and Tamil).
- Implementation of transformer-based models (mBERT, IndicBERT) for data in both languages (Malayalam and Tamil)
- Performance evaluation of the different models implemented on a standard dataset using standard metric.

The rest of the paper is organized as follows: Section 2 discusses the related work for the detection of homophobia and transphobia. Section 3 describes the dataset and its dimensions. Section 4 presents the experimental set up. Results are presented in Section 5. Section 6 concludes the paper with a brief summary of the results and major conclusions drawn.

## 2 Related work

There is very limited previous work on detecting homophobia and transphobia in social media comments, particularly on low resource Indian languages. In one study, the ensemble transformer-based model was implemented to classify homophobia and transphobia on social media comments in Tamil and Tamil-English datasets [1]. Similarly, to classify the comments as homophobia and transphobia in English, different transformer-based models (BERT, RoBERTa and HateBERT) were implemented [2]. For detecting homophobia and transphobia, different monolingual and multilingual transformer-based models experimented on English, Tamil and Tamil-English datasets [3]. Another study experimented with transformer-based model (RoBERT-base) for English, Tamil and English-Tamil code-mixed datasets [6]. A study implemented neural network using sentence embedding and ensemble model in Tamil, English and Tamil-English [4]. Some of the studies implemented different classification algorithms using TF-IDF to classify the YouTube comments as homophobic and transphobic in English, Tamil and Tamil-English [5]. Similarly, the combination of word embeddings and Support Vector Machine (SVM) are implemented along with BERT-based models in English, Tamil and English-Tamil [7]. There is, however, no previous work on detecting homophobia and transphobia on social media comments in Malayalam language. Further, the previous studies have not explored the newer methods like IndicBERT. Therefore, the present study explores the suitability of various deep learning-based models, including the recently proposed IndicBERT model, for the task of detecting homophobia and transphobia on social media comments in Malayalam language Tamil languages.

# 3 Dataset Description

The dataset used in the study was launched by DravidianLangTech and is available online at https://codalab.lisn.upsaclay.fr/competitions/. The dataset contains the comments in two Dravidian languages namely Malayalam and Tamil. There are two attributes in the dataset: comment and category. The comment attribute consists of the sentiment/opinion of the people in form of text and the category tells that which of the three classes the comment belongs to: Homophobic, Transphobic, and Non-anti-LGBT+content. The Malayalam dataset contains 3,114 comments and the Tamil dataset contains 2,662 comments. Table 1 shows the category-wise distribution of comments in the Malayalam dataset. Table 2 shows the distribution of comments in each category of the Tamil dataset.

**Table 1.** Category-wise distribution of comments of Malayalam Dataset

| Category | No. of comments |
|---|---|
| Homophobic | 2434 |
| Transphobic | 491 |
| Non-anti-LGBT+content | 189 |

**Table 2.** Category-wise distribution of comments in Tamil Dataset

| Category | No. of comments |
|---|---|
| Homophobic | 2022 |
| Transphobic | 485 |
| Non-anti-LGBT+content | 155 |

## 3.1 Data pre-processing

Since data pre-processing is a very important step in NLP applications and model building, we performed the following steps to pre-process the original text:

- Removing punctuations
- Removing Numbers
- URLs given in comments are removed.
- Repeated characters that do not add any information are removed.

- Extra blank spaces, that are created during pre-processing or if some comments already have that, are removed.

# 4 Experimental Setup

## 4.1 CNN

CNN was initially developed for image processing but in the past years, CNN showed good results in NLP tasks as well. When CNN is applied to text rather than the image, then 1-dimensional convolutional neural networks are used. To work with 1-D CNN, a word embedding layer is required. GLoVe word embedding is used in this paper. For preparing the word embedding for the model, firstly all punctuations, links, etc. are removed, the text is tokenized and padding the text using the pad_sequences() to make equal length sequences. Fig.1 shows the implementation of CNN. From the word embedding, the words are represented as vectors. The embedding matrices are passed to the embedding layer. After that, the convolutional layer is added using “relu” as an activation function. Now, a pooling layer is added to reduce the complexity of the dimensions, and to be more specific MaxPooling is used. At last, the fully connected layer is added which is also called the dense layer in which the “Softmax” is used as an activation function since the dataset is multi-class classification. So, it classifies the comments into three categories: Homophobic, Transphobic and Non-anti-LGBT+content. The model is trained on 100 epochs for both languages (Malayalam and Tamil) and “Adam” is used as an optimizer. The loss is” Categorical Cross-entropy” to classify the tweets.

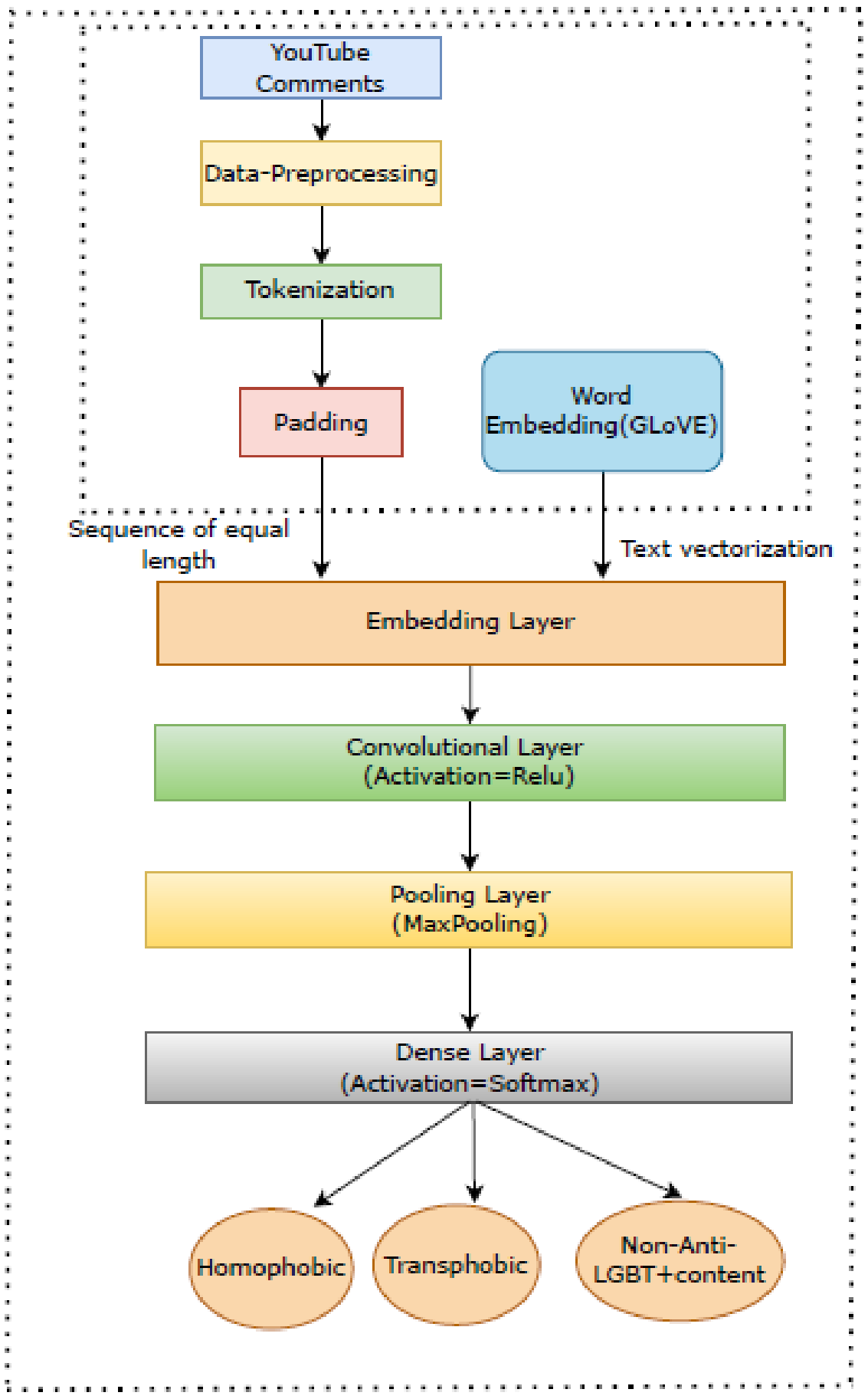

**Fig. 1.** Implementation of CNN to detect homophobia and transphobia

### 4.2 LSTM

LSTM is used for text classification because LSTM memorizes the information effectively and removes unwanted information. The LSTM model is able to find out the meaning of the input string and classify the text to the most precise output class if the appropriate layers of embedding are used. Before adding the embedding layer, firstly the data is pre-processed and the text is tokenized into tokens. After that, the sequence is padded by adding zero at the end of the sequence to make the sequence of the same length. Now, the word embedding (GLoVe) is used to represent the words into vectors. The padded sequence are passed to embedding layer as an input. The LSTM layer has been added after embedding layer using ReLu as an activation function. The dense layer is added at the end with activation function as softmax to classify the comments into more than two classes. Fig.2 shows the block diagram for LSTM to classify the comments. The model is trained on 100 epochs using the categorical cross-entropy as a loss function and "Adam" as an optimizer for both languages.

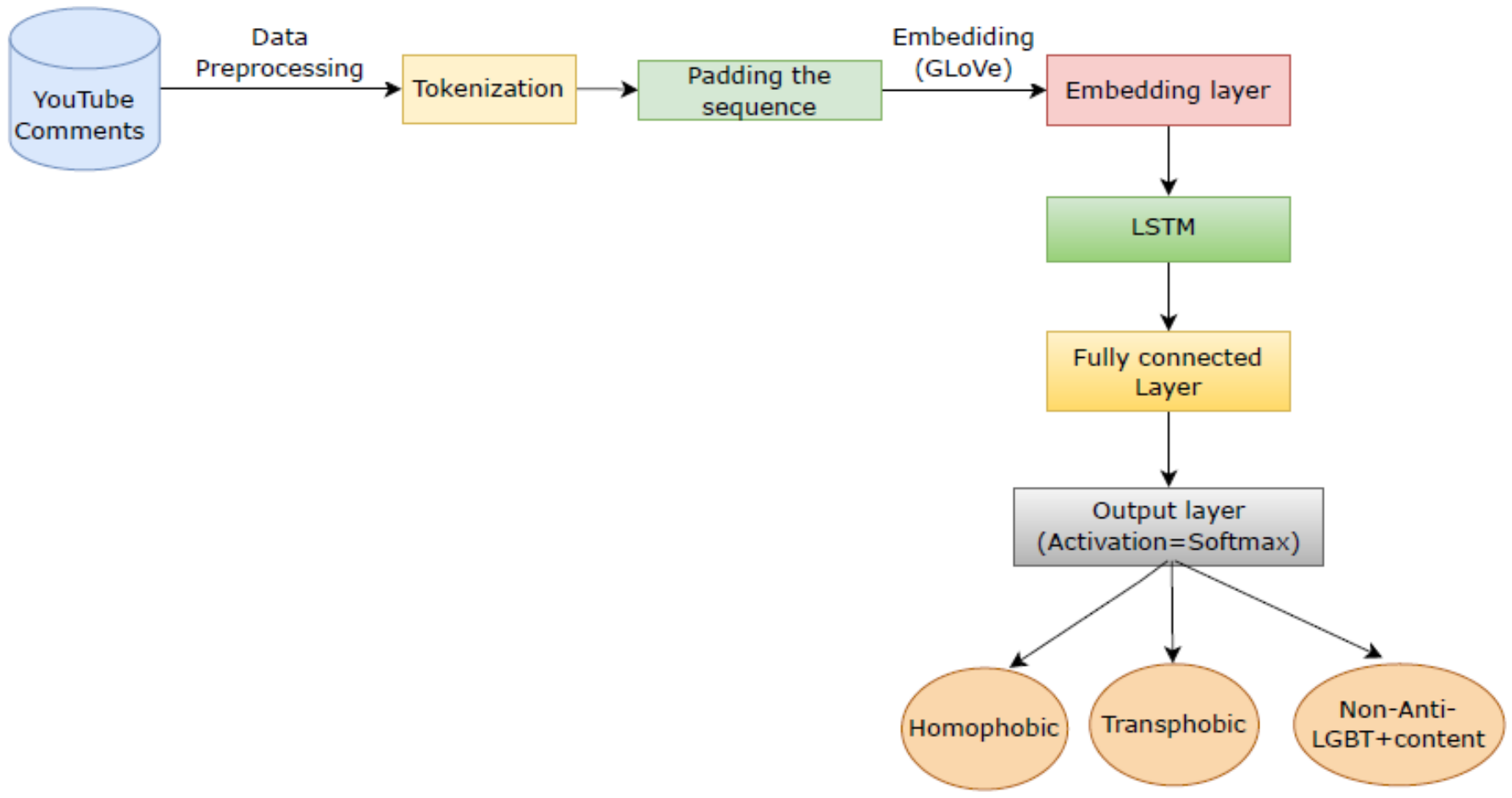

**Fig. 2.** Block diagram to show the implementation of LSTM model

## 4.3 mBERT

BERT was developed in 2018 by Jacob Devlin and his colleagues from Google. BERT was trained on large-corpus of English data whereas mBERT was trained on 104 different languages including English [19]. It is not feasible to have separate BERT model for each language and therefore mBERT was introduced as a single BERT for 104 languages. In this paper, BERT base is used that has 12 layers of transformer encoder. Firstly, the comments are converted into tokens with the help of BERT tokenizer. Fig.3 shows the framework of BERT to classify the comments. Two special tokens [CLS] and [SEP] are added at the starting and end of the sequence, respectively. Vocabulary IDs for each of the tokens are token embeddings. Sentence embedding is used to distinguish between the sentences; and positional embedding is used to show the position of each word in the sequence. The model is build using the pre-trained BERT-base multilingual-cased model. The batch size is 32, learning is 3e-5 and the epochs=5 is used to train the model on the dataset. At last the classifier classify the comments into homophobic, transphobic and non-anti-LGBT+content classes.

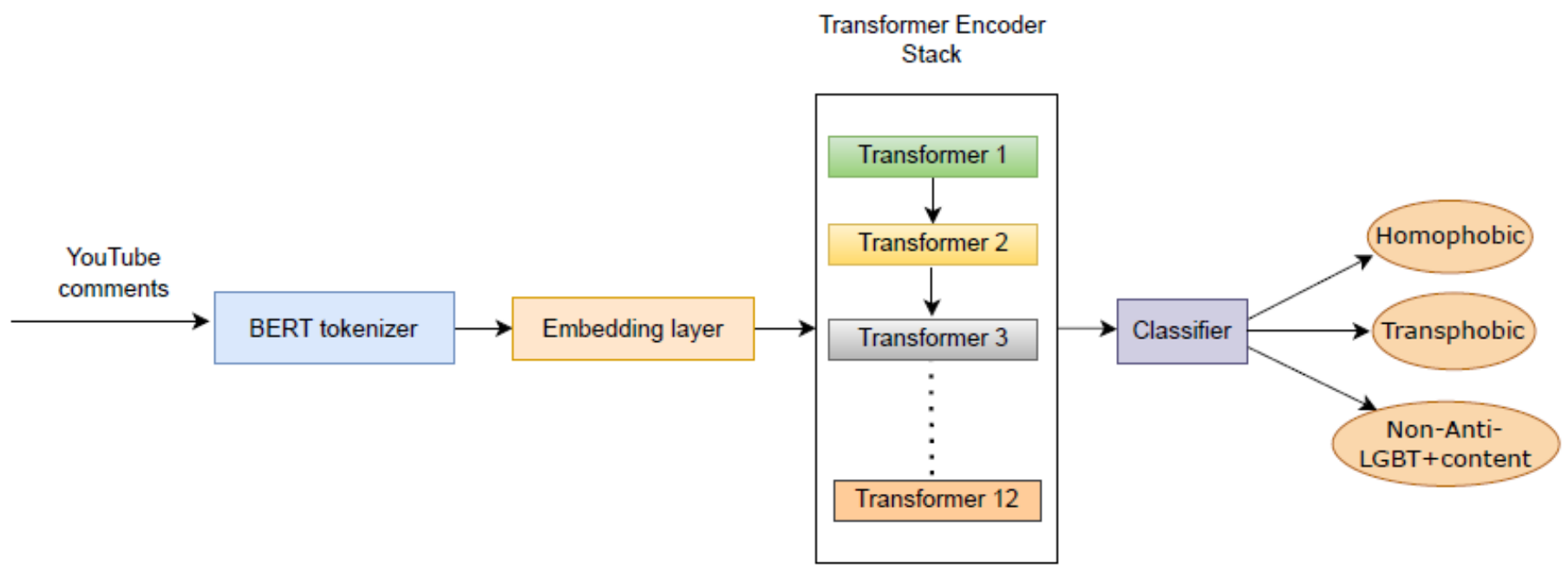

**Fig. 3.** Framework of transformer-based BERT model

### 4.4 IndicBERT

IndicBERT is a multilingual model trained on 12 Indian languages: Tamil, Malayalam, Bengali, Marathi, Punjabi, English, Hindi, Gujarati, Assamese, Oriya, Telugu and Kannada. As compared to other models like mBERT and XLM-R, it has a smaller number of parameters but even then gives the state-of-the-art (SOTA) performance for various tasks [18]. IndicBERT is based on the ALBERT model which is the derivative of the BERT. The text is tokenized using the Autotokenizer and the [CLS] token is added at the beginning of the sequence and [SEP] tokens are added at the end of the sequence similar to the BERT model. IndicBERT also has the ability to consider more than one sentence into a single sequence for input. As the dataset is in Indian languages: Malayalam and Tamil. For this dataset, the IndicBERT outperformed the other models and reported the weighted average F1 score for Malayalam and Tamil is 0.86 and 0.77, respectively. The parameters are fine-tuned: learning rate is set to be 3e-5, epochs=5 and batch size is 32. At the end, the model classifies the comments into the three given categories: Homophobic, Transphobic and non-anti-LGBT+content.

## 5 Results

The CNN and transformer-based models are implemented to classify the comments from two languages into one of the three classes: ‘Homophobic’, ‘Transphobic’, and ‘Non-anti-LGBT+content’. This section shows the performance of the implemented models. Table 3 shows the performance analysis of the CNN, LSTM and transfer learning models (mBERT and IndicBERT) for Malayalam. It can be observed that the weighted F1-score for CNN, LSTM, mBERT and IndicBERT are 0.80, 0.81, 0.83 and 0.86, respectively. For the Malayalam language, IndicBERT outperformed the other models. Table 4 shows the performance analysis of the implemented models for Tamil. The weighted average F1-score for deep learning models: CNN and LSTM are 0.71 and 0.72, respectively. The transformer-based model (IndicBERT) performs better than the other three implemented models (CNN, LSTM, mBERT), with a weighted F1 score of 0.77.

**Table 3.** Performance analysis for implemented models for Malayalam Dataset

| Model | Homophobic | | | Transphobic | | | Non-anti-LGBT+content | | | Weighted Average F1-score |
|---|---|---|---|---|---|---|---|---|---|---|
| | P | R | F1-score | P | R | F1-score | P | R | F1-score | |
| CNN(GloVe) | 0.65 | 0.35 | 0.45 | 0.86 | 0.32 | 0.46 | 0.84 | 0.96 | 0.89 | 0.80 |
| LSTM(GloVe) | 0.69 | 0.45 | 0.54 | 0.60 | 0.32 | 0.41 | 0.86 | 0.95 | 0.90 | 0.81 |
| mBERT | 0.74 | 0.47 | 0.58 | 0.67 | 0.32 | 0.43 | 0.86 | 0.96 | 0.91 | 0.83 |
| IndicBERT | 0.79 | 0.49 | 0.59 | 0.70 | 0.39 | 0.50 | 0.88 | 0.97 | 0.91 | **0.86** |

*P=Precision, R=Recall

**Table 4.** Performance analysis for implemented models for Tamil Dataset

| Model | Homophobic | | | Transphobic | | | Non-anti-LGBT+content | | | Weighted Average F1-score |
|---|---|---|---|---|---|---|---|---|---|---|
| | P | R | F1-score | P | R | F1-score | P | R | F1-score | |
| CNN(GloVe) | 0.57 | 0.08 | 0.14 | 1.00 | 0.27 | 0.42 | 0.78 | 0.99 | 0.88 | 0.71 |
| LSTM(GloVe) | 0.67 | 0.12 | 0.21 | 0.80 | 0.27 | 0.40 | 0.79 | 0.98 | 0.87 | 0.72 |
| mBERT | 0.70 | 0.15 | 0.24 | 0.79 | 0.30 | 0.43 | 0.81 | 0.96 | 0.87 | 0.74 |
| IndicBERT | 0.72 | 0.18 | 0.28 | 0.76 | 0.34 | 0.46 | 0.84 | 0.94 | 0.88 | **0.77** |

*P=Precision, R=Recall

## 6 Conclusion

The paper presents the performance analysis of the different deep learning models (CNN, LSTM, mBERT and IndicBERT) on the task of homophobia and transphobia detection of the Malayalam and Tamil texts. The goal was to classify each comment into given three categories: Homophobic, Transphobic and Non-anti-LGBT+content. Experimental results show that the IndicBERT model outperformed the other models for both languages: Malayalam and Tamil, with reported weighted average F1-score of 0.86 and 0.77, respectively. While the other models also show reasonable performance, but the IndicBERT model is the best among all in terms of measured performance values. Thus, the IndicBERT model emerges as the superior of the various proposed methods for the tasks of Homophobia and Transphobia detection in the selected Dravidian languages. The present work can also be extended to data from other Indian languages. Further, ensemble-based models can be explored for the given task.